\documentclass[conference]{IEEEtran}
\IEEEoverridecommandlockouts
\usepackage{cite}
\usepackage{amsmath,amssymb,amsfonts}
\usepackage{algorithmic}
\usepackage{graphicx}
\usepackage{textcomp}
\usepackage{xcolor}
\usepackage[nolist]{acronym}
\usepackage{cite}   % for compact citations, i.e. [1, 2]
\usepackage{caption}  % Ensure this is included for normal captioning

\usepackage{subcaption}  % For creating subtables
\usepackage{tabularx}    % For flexible table widths
\usepackage{booktabs}    % For better table lines
\usepackage{multirow}    % For multi-row cells
\usepackage{balance} % for balanced references across two columns on the last page
\usepackage{bm}
\usepackage{pgfplots}
\usepackage{pgfplotstable}
\usepackage{siunitx} % for units of meaurements
\usepackage[colorlinks=true,allcolors=blue]{hyperref} % hyperlink
\usepackage{balance}   %

\usetikzlibrary{matrix}

\usepackage{listings}
\makeatletter
\def\@citex[#1]#2{\leavevmode
\let\@citea\@empty
\@cite{\@for\@citeb:=#2\do
{\@citea\def\@citea{,\penalty\@m\ }%
\edef\@citeb{\expandafter\@firstofone\@citeb\@empty}%
\if@filesw\immediate\write\@auxout{\string\citation{\@citeb}}\fi
\@ifundefined{b@\@citeb}{\hbox{\reset@font\bfseries ?}%
\G@refundefinedtrue
\@latex@warning
{Citation `\@citeb' on page \thepage \space undefined}}%
{\@cite@ofmt{\csname b@\@citeb\endcsname}}}}{#1}}
\makeatother

\begin{acronym} 
    \acro{PINC}{physics informed neural network with control}
    \acro{PINN}{physics informed neural network}
    \acro{DMD}{dynamic mode decomposition}
    \acro{SINDyC}{sparse identification of non-linear dynamics with control}
   \acro{DMDc)}{ dynamic mode decomposition with control}
    \acro{MPC}{model predictive control}
    \acro{GP}{Gaussian process}
    \acroplural{GP}[GPs]{Gaussian processes}
    \acro{AUV}{Autonomous underwater vehicle}
    \acro{EKF}{extended kalman filter}
    \acro{ROV}{remotely operated vehicle}
    \acro{IMU}{inertial measurement unit}
    \acro{DVL}{doppler velocity log}
    \acro{NED}{North-EastDown}
    \acro{NN}{neural network}
\end{acronym}

\newtheorem{remark}{Remark}

\def\BibTeX{{\rm B\kern-.05em{\sc i\kern-.025em b}\kern-.08em
    T\kern-.1667em\lower.7ex\hbox{E}\kern-.125emX}}

\begin{document}

\title{Modelling of Underwater Vehicles using Physics-Informed Neural Networks with Control
\thanks{This research was partially supported by the Aarhus University Research Foundation, EIVA a/s and Innovation Fund Denmark under grant 2040-00032B.}
\thanks{$^{*}$These authors contributed equally to this work.}
}

\author{%
\IEEEauthorblockN{Abdelhakim Amer$^{*}$}
\IEEEauthorblockA{\textit{Department of Electrical and Computer Engineering} \\
\textit{Aarhus University}\\
Aarhus, Denmark \\
abdelhakim@ece.au.dk}
\and
\IEEEauthorblockN{David Felsager$^{*}$}
\IEEEauthorblockA{\textit{Department of Electrical and Computer Engineering} \\
\textit{Aarhus University}\\
Aarhus, Denmark \\
201904960@post.au.dk}
\and
\IEEEauthorblockN{Yury Brodskiy}
\IEEEauthorblockA{\textit{EIVA a/s} \\
Skanderborg, Denmark \\
ybr@eiva.com}
\and
\IEEEauthorblockN{Andriy Sarabakha}
\IEEEauthorblockA{\textit{Department of Electrical and Computer Engineering} \\
\textit{Aarhus University}\\
Aarhus, Denmark \\
andriy@ece.au.dk}
}

% \author{\IEEEauthorblockN{Anonymous Authors}}

\maketitle

\begin{abstract}

Physics-informed neural networks~(PINNs) integrate physical laws with data-driven models to improve generalization and sample efficiency. This work introduces an open-source implementation of the Physics-Informed Neural Network with Control (PINC) framework, designed to model the dynamics of an underwater vehicle. Using initial states, control actions, and time inputs, PINC extends PINNs to enable physically consistent transitions beyond the training domain. Various PINC configurations are tested, including differing loss functions, gradient-weighting schemes, and hyperparameters. Validation on a simulated underwater vehicle demonstrates more accurate long-horizon predictions compared to a non-physics-informed baseline. 

\end{abstract}

\begin{IEEEkeywords}
physics-informed machine learning, dynamics modelling, underwater robotics
\end{IEEEkeywords}

%%%%%%%%%%%%%%%%%%%%%%%%%%%%%%%%%%%%%%%%%%%%%%%%%%%%%%%%%%%%%%%%%%%%%%%%%%%%%%%%%%%%%%%%%%%%%%%%%%%%
%%%%%%%%%%%%%%%%%%%%%%%%%%%%%%%%%%%%%%%%%%%%%%%%%%%%%%%%%%%%%%%%%%%%%%%%%%%%%%%%%%%%%%%%%%%%%%%%%%%%
%%%%%%%%%%%%%%%%%%%%%%%%%%%%%%%%%%%%%%%%%%%%%%%%%%%%%%%%%%%%%%%%%%%%%%%%%%%%%%%%%%%%%%%%%%%%%%%%%%%%

\section{Introduction} % <1 page
% Motivation
% Relevance of ROVs
Underwater robotic systems, such as autonomous underwater vehicles (AUVs) and remotely operated vehicles (ROVs), are critical for tasks such as seabed inspections, pipeline monitoring \cite{amer2023unav}, and deep-sea exploration \cite{kunz2008deep}, where human access is limited. With advances in autonomy, underwater robots can tackle complex manipulation tasks, including pipeline repair and maintenance. 
%
% Importance of modelling for control
%
Such applications require precise control and accurate dynamic models. For example, model predictive control (MPC) provides a framework for achieving complex tasks with high performance, such as inspection \cite{amer2023visual}, while relying on an accurate motion model for prediction. However, modelling underwater robots using first principles is challenging due to non-linearities arising from hydrodynamic disturbances, and unmodeled higher-order effects \cite{lakshminarayanan2024estimation}. Furthermore, due to the wide variety of underwater vehicle designs, sizes, and degrees of freedom, modeling is often cumbersome and, in some cases, impractical. Data-driven methods on the other-hand offer a unifying framework to estimate the dynamics of these diverse systems.
%
% Justification for need of PINN and PINC
Physics-informed neural networks~(PINNs) in particular is a promising approach that integrates data-driven neural networks with physical laws as regularization to produce physically plausible outputs \cite{karniadakis2021physics,drgona2025safe}. 

%
% Contribution
This work explores the potential of a specialized variant of PINNs, namely the physics-informed neural network with control~(PINC) \cite{antonelo_physics-informed_2024}, for modelling the dynamics of underwater ROVs. The primary objective is to evaluate whether PINC can effectively model a simplified underwater dynamic system and have the potential to provide an accurate model for control applications.
The investigation applies these modelling approaches to an underwater ROV as the physical system of interest. While prior studies on PINC have demonstrated potential benefits in other domains, no existing work (as of this writing) appears to apply PINC and its implicit integration technique, combined with autoregressive predictions, to underwater robotics. This gap presents a unique opportunity to evaluate whether and how PINC might offer advantages -- such as improved accuracy or generalization -- over a traditional DNN when modelling complex underwater dynamics. To that end, the model’s predictive capabilities and robustness are tested by incorporating input noise during training, among other methods. The main contributions of this article are summarized as follows:

\begin{itemize}
    \item Developed a PINC framework to model simplified ROV dynamics, achieving high long-horizon prediction accuracy with minimal computational overhead.
    \item Demonstrated the effectiveness of combining physics-based regularization with training optimizations, including residual connections, gradient normalization, and adaptive learning rate scheduling.
    \item Performed a comprehensive analysis of the hyperparameters' effect on the PINC performance.
    \item Released an open-source implementation of the PINC framework\footnote{\url{https://github.com/eivacom/pinc-xyz-yaw}}, including a simulation model of an underwater vehicle for synthetic data generation.
   % \item Made available a real-world dataset, collected using a motion capture system for a BlueROV, enabling researchers to train and evaluate their own PINC architectures.
\end{itemize}

%\end{itemize}

% Organisation
This paper is organised as follows. Section~\ref{sec:related} provides a comprehensive literature review on PINNs, including their applications and limitations. Section~\ref{sec:problem} motivates the importance of accurate and computationally efficient dynamics models for enhancing prediction and control. Section~\ref{sec:method} outlines the methodological framework for evaluating PINC and baseline models in simulation experiments. Section~\ref{sec:results} presents the simulation results for different PINC configurations and baseline comparisons. Finally, Section~\ref{sec:conclusions} concludes with a summary of key insights and suggestions for future research directions.

%%%%%%%%%%%%%%%%%%%%%%%%%%%%%%%%%%%%%%%%%%%%%%%%%%%%%%%%%%%%%%%%%%%%%%%%%%%%%%%%%%%%%%%%%%%%%%%%%%%%
%%%%%%%%%%%%%%%%%%%%%%%%%%%%%%%%%%%%%%%%%%%%%%%%%%%%%%%%%%%%%%%%%%%%%%%%%%%%%%%%%%%%%%%%%%%%%%%%%%%%
%%%%%%%%%%%%%%%%%%%%%%%%%%%%%%%%%%%%%%%%%%%%%%%%%%%%%%%%%%%%%%%%%%%%%%%%%%%%%%%%%%%%%%%%%%%%%%%%%%%%

\section{Related Works} % ~1 page
\label{sec:related}
Traditional data-driven methods for non-linear system identification and control range from flexible but opaque neural networks to interpretable but potentially limited approaches like dynamic mode decomposition with control~(DMDc). Sparse identification of non-linear dynamics with control~(SINDyC)~\cite{brunton_data-driven_2021} constructs a large function library, then applies a sparsity-promoting algorithm to isolate key dynamics for model-based control. Gaussian processes~(GPs) are widely used for data-driven modeling~\cite{amer2025} due to their ability to quantify uncertainty, which is advantageous for designing safe control architectures~\cite{liang2025adaptive}. However, they often scale poorly in high-dimensional spaces.  PINNs address these trade-offs by embedding physical laws (differential equations) directly into the neural network's loss, enhancing data efficiency and physical consistency within a flexible framework.

%\textbf{PINN Variations.} 
Many works extend PINNs for more complex tasks. For instance, \cite{faria_data-driven_2024} applies PINNs to Reinforcement Learning, yielding faster training and robust controllers beyond the training domain. \cite{ramp} introduces a Robust Adaptive MPC framework leveraging PINNs to integrate physics-based priors with data-driven learning, improving trajectory tracking performance under uncertainties while mitigating the computational burden of traditional robust MPC approaches.

Hybrid architectures like \textit{Deep Lagrangian networks}~\cite{lutter_combining_2023} embed energy-based formulations for better interpretability. Others focus on \textit{Neural ODEs}~\cite{ma_development_2024, chen2018neural}, augmenting nominal physics models with learned residuals. PINNs have also found success modelling AUV dynamics~\cite{zhao_research_2024}, soft robotics~\cite{gao_sim--real_2024}, and more. PINC extends PINNs to include control inputs, initial states, and time~\cite{antonelo_physics-informed_2024}. PINC can model a continuum of initial conditions by treating the network as a continuous time-stepper. Auto-regressive rollouts allow multi-step prediction, and architectures like domain-decoupled PINNs~\cite{krauss_domain-decoupled_2024} can boost training efficiency. PINC has shown promise for efficient MPC integration and large-scale industrial tasks~\cite{kittelsen_physics-informed_2024}. Gradient scaling poses a significant challenge when dealing with multiple loss terms, such as data, physics, and rollout losses, which may conflict with each other. Techniques like ConFIG \cite{liu_config_2024} address this issue by aligning gradient directions to mitigate conflicts. Additionally, activation smoothness plays a critical role in PINN performance. As highlighted in \cite{nicodemus_physics-informed_2022}, the use of non-smooth activation functions can impair performance due to the inaccuracies introduced by automatic differentiation.
%Gradient scaling is a key hurdle multiple loss terms (data, physics, and rollout) can conflict, motivating techniques like ConFIG (Conflict-free Training of PhysicsInformed Neural Networks) \cite{liu_config_2024} to align gradient directions. Activation smoothness \cite{nicodemus_physics-informed_2022} is also crucial since automatic differentiation of non-smooth activations can degrade PINN performance. 

%%%%%%%%%%%%%%%%%%%%%%%%%%%%%%%%%%%%%%%%%%%%%%%%%%%%%%%%%%%%%%%%%%%%%%%%%%%%%%%%%%%%%%%%%%%%%%%%%%%%
%%%%%%%%%%%%%%%%%%%%%%%%%%%%%%%%%%%%%%%%%%%%%%%%%%%%%%%%%%%%%%%%%%%%%%%%%%%%%%%%%%%%%%%%%%%%%%%%%%%%
%%%%%%%%%%%%%%%%%%%%%%%%%%%%%%%%%%%%%%%%%%%%%%%%%%%%%%%%%%%%%%%%%%%%%%%%%%%%%%%%%%%%%%%%%%%%%%%%%%%%

\section{Problem Formulation} % ~1 page
\label{sec:problem}

PINNs are effective for solving nonlinear dynamical systems by leveraging both partial knowledge of system dynamics and available data. For a general nonlinear system, where the system’s dynamics are represented by a function that is Lipschitz continuous, the solution over time depends on the complete knowledge of this function. However, in practice, this function is often not fully known due to uncertainties and modelling inaccuracies. Instead, nominal dynamics are used, and PINNs approximate the solution using a neural network, which incorporates these nominal dynamics into its training process. %This approach improves the efficiency of data usage and provides better robustness in the presence of uncertainties compared to traditional data-driven methods, which typically rely on large labelled datasets.
\begin{figure}[b!]
    \centering
    \includegraphics[width=0.95\columnwidth]{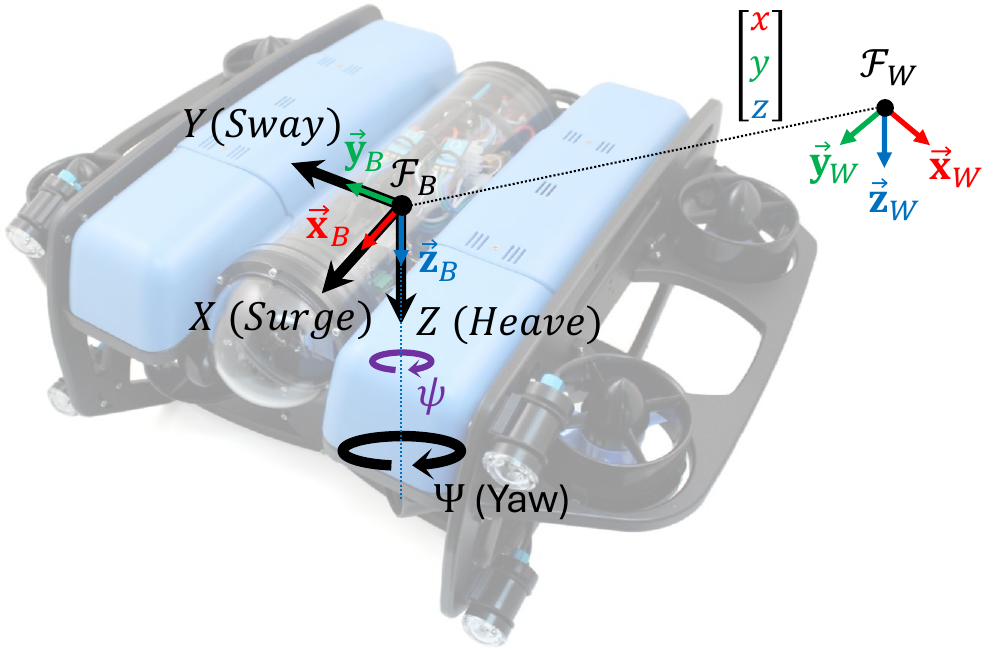}
    \caption{Coordinate frame of the ROV with respect to the world frame.}
    \label{fig:rov-coord-frame}
\end{figure}
In this work, PINNs are applied to the open-loop system identification of the BlueROV2, a versatile underwater ROV commonly used for inspection and manipulation tasks. The BlueROV2 and its coordinate frame are illustrated in Fig.~\ref{fig:rov-coord-frame}, where it is noted that the vehicle has six thrusters, which provide input commands in the form of forces and torques around the ROV's center of gravity, with the thruster dynamics excluded from the model. A simplified 4-degree-of-freedom version of the “Fossen” model ~\cite{fossen}, with assumptions of small pitch and roll, is adopted for modelling the nominal system dynamics, which can be expressed in state-space form as follows:
\begin{equation}
    \label{eq:rov_simplified}
    \begin{aligned}
        \dot{\mathbf{x}} = \begin{bmatrix}
        \cos(\psi)u-\sin(\psi)v \\ 
        \sin(\psi)u+\cos(\psi)v \\ 
        w \\ 
        r \\ 
        \frac{1}{m-X_{\dot{u}}}\left(X + (m-Y_{\dot{v}})vr+(X_u+X_{u|u|}|u|)u\right) \\ 
        \frac{1}{m-Y_{\dot{v}}}\left(Y - (m-X_{\dot{u}})ur+(Y_v+Y_{v|v|}|v|)v\right) \\ 
        \frac{1}{m-Z_{\dot{w}}}\left(Z + (Z_w+Z_{w|w|}|w|)w+F_g-V_{sub } \rho_{water}\right) \\ 
        \frac{1}{I_{zz}-N_{\dot{r}}}\left(\Psi - (X_{\dot{u}}-Y_{\dot{v}})uv+(N_r+N_{r|r|}|r|)r\right) 
    \end{bmatrix},
    \end{aligned}
\end{equation}
The state vector is defined as $\mathbf{x} = [x, y, z, \psi, u, v, w, r]$, where $x$, $y$, and $z$ represent the translational positions in the world-fixed frame $\mathcal{F}_W$, and $u$, $v$, and $w$ denote the translational velocities in the body-fixed frame $\mathcal{F}_B$. The transformation of translational velocities from the body-fixed frame $\mathcal{F}_B$ to the world-fixed frame $\mathcal{F}_W$ is achieved by rotating the body velocities $(u, v)$ around $z_b$ using the yaw angle $\psi$. Additionally, the rotational velocity is represented by $r$. The input vector, defined as $\mathbf{u} = [X, Y, Z, \Psi]$, comprises the forces $X$, $Y$, and $Z$, which are the linear components of the combined force applied by the actuators on the AUV body in $\mathcal{F}_B$, and $\Psi$, which denotes the actuation control moment around $z_b$. 
The vehicle’s physical parameters include its mass $m$, the angular moment of inertia about the z-axis $I_{zz}$, the gravitational constant $g$, the water density $\rho_{water}$, and the total submerged volume $V_{sub}$. The added mass coefficients, are given by $X_{\dot{u}}$, $Y_{\dot{v}}$, $Z_{\dot{w}}$, and $N_{\dot{r}}$. Drag effects are characterized by linear coefficients $X_{u}$, $Y_{v}$, $Z_{w}$, and $N_{r}$, and quadratic drag coefficients $X_{uc}$, $Y_{vc}$, $Z_{wc}$, and $N_{rc}$, which describe the drag along the surge, sway, heave, and yaw directions, respectively.

%where $\mathbf{x} = [x,\, y,\, z,\, \psi,\, u,\, v,\, w,\, r]^\top$ denotes pose and velocities, and $\mathbf{u} = [X,\, Y,\, Z,\, \Psi]^\top$ are control forces/torques.  Non-linear damping, buoyancy terms, and rotational dynamics lead to complex system behaviour.

%%%%%%%%%%%%%%%%%%%%%%%%%%%%%%%%%%%%%%%%%%%%%%%%%%%%%%%%%%%%%%%%%%%%%%%%%%%%%%%%%%%%%%%%%%%%%%%%%%%%
%%%%%%%%%%%%%%%%%%%%%%%%%%%%%%%%%%%%%%%%%%%%%%%%%%%%%%%%%%%%%%%%%%%%%%%%%%%%%%%%%%%%%%%%%%%%%%%%%%%%
%%%%%%%%%%%%%%%%%%%%%%%%%%%%%%%%%%%%%%%%%%%%%%%%%%%%%%%%%%%%%%%%%%%%%%%%%%%%%%%%%%%%%%%%%%%%%%%%%%%%

\section{Proposed Method} % ~1 pages
\label{sec:method}

\subsection{PINC: Physics-Informed Neural Network with Control}

PINNs \cite{raissi_physics-informed_2019} incorporate known dynamics into a neural network via a \emph{physics loss} enforcing consistency of \eqref{eq:rov_simplified}. Standard PINNs take continuous time $t$ and the state $\mathbf{x}(0)$ as inputs. 
PINC \cite{antonelo_physics-informed_2024} extends PINNs to handle control inputs. We model the solution $\mathbf{x}(t)$ given:
\begin{equation}
\hat{\mathbf{x}}(t) = \mathcal{N}\Bigl(\left[\,\mathbf{x}(0),\,\mathbf{u}(0),\,t\right]\Bigr),
\quad \text{for } t\in[\,0,T\,],
\label{eq:pinc_mapping}
\end{equation}
where $\mathbf{u}(0)$ is the control at the initial step, assumed constant in $[0,T]$. %PINC can be iterated autoregressively to predict $\mathbf{x}(kT)$ for $k=1,2,\dots$.

Discrete-time steps are denoted by index $n$, and continuous time by $t$. The $n$th initial state in a trajectory is $\mathbf{x}_n(0)$ with corresponding control actions $\mathbf{u}_n(0)$. Each point in a trajectory serves as an initial condition, making the one-step-ahead prediction $\hat{\mathbf{x}}_n(T)$ approximate the next state $\mathbf{x}_{n+1}(0)$:
\begin{equation}
    \mathbf{x}_{n}(T) = \mathbf{x}_{n+1}(0).
\end{equation}
Thus, the one-step-ahead predicted state is:
\begin{equation} 
    \hat{\mathbf{x}}_{n}(T) 
    =\mathcal{N}\bigl(\underbrace{\begin{bmatrix}
    \mathbf{x}_n(0) & \mathbf{u}_n(0) & T
\end{bmatrix}}_{\mathbf{z}_n(0)}\bigr)
    \approx\mathbf{x}_{n+1}(0).
\end{equation}
Long-horizon predictions (rollouts) over $N$ steps are defined as:
\begin{equation}
    \mathbf{x}_0(NT)=\mathbf{x}_N(0).
\end{equation}
These are achieved by autoregressively applying the dynamics model $\mathcal{N}$:
\begin{equation}
\label{eq:rollout_prediction}
    \hat{\mathbf{x}}_0(NT) = \underbrace{\mathcal{N} \Bigl( \mathcal{N} \bigl( \mathbf{x}_0(0), \mathbf{u}_0(0), T \bigr)\dots, \mathbf{u}_{N-1}(0), T\Bigr)}_{\text{N times}}.
\end{equation}
During training, multiple trajectories are batched together indexed by $m$ (batch size $N_B$), each containing $N_D$ points indexed by $n$. For multi-step predictions and physics loss computations, the index $k$ and number of points $N_{P}$ are respectively used for the number of prediction steps and number of physics collocation points. 

\subsection{Neural Network Architecture}
\label{sec:nn_arch}
The residual deep neural network architecture is used to learn ROV dynamics in a physics-informed manner.
\subsubsection{Residual Formulation for ODE Integration}
Solving an ODE over a time interval $I = [0,T]$ can be written as
\begin{equation}
    \mathbf{x}(T) = \mathbf{x}(0) + \int_0^T f(\mathbf{x}(\tau), \mathbf{u}(\tau)) d\tau.
\end{equation}
When the control is assumed constant over $I$ (zero-order hold), the integral can be approximated with a neural network that learns to “integrate” the dynamics from $\mathbf{x}(0)$ to $\mathbf{x}(T)$:
\begin{equation}
    \mathbf{x}(T) \approx \mathbf{x}(0) + \mathcal{N}\bigl(\underbrace{\begin{bmatrix}
    \mathbf{x}(0) & \mathbf{u}(0) & T
\end{bmatrix}}_{\mathbf{z}(0)}\bigr).
\end{equation}

\subsubsection{Layers, Neurons, and Activations}
The architecture includes fully connected $N_L$ hidden layers, each containing $N_H$ neurons. Both adaptive \lstinline|tanh| and \lstinline|softplus| activation functions are tested with an adaptable parameter, denoted $\beta$, that is unique for each layer.

\subsubsection{State Re-Parameterization for Yaw}
Since yaw angle $\psi$ wraps around at $\pm\pi$, it is replaced with two states: $\cos(\psi)$ and $\sin(\psi)$. This ensures the continuity of the states and avoids angle discontinuities. 

\subsubsection{Rotational Structural Information}
To reflect the natural geometry of planar motion, the network’s predicted increments in $x$ and $y$ are rotated from the body frame $\mathcal{F}_B$ to the world-fixed $\mathcal{F}_W$. Specifically, if the raw network outputs are $\Delta \hat{x}_b$ and $\Delta \hat{y}_b$, they are transformed as follows:
\begin{equation}
    \begin{bmatrix}
        \Delta\hat{x}_n\\
        \Delta\hat{y}_n
    \end{bmatrix}=
    \begin{bmatrix}
        \cos(\hat{\psi}) & -\sin(\hat{\psi})\\
        \sin(\hat{\psi}) & \cos(\hat{\psi})
    \end{bmatrix}
    \begin{bmatrix}
        \Delta\hat{x}_b\\
        \Delta\hat{y}_b
    \end{bmatrix}.
\end{equation}
Because the model learns increments in the body frame, it can capture the simpler local dynamics. Those increments are then rotated back into $\mathcal{F}_W$.

\subsubsection{Layer Normalization} 
Layer normalization \cite{ba_layer_2016} is another regularization technique employed in this work, which normalizes activations within each layer using learnable parameters, mitigating internal covariate shifts. 

\subsection{Loss Functions}
\label{sec:loss_funcs}
PINC integrates multiple loss terms to accurately learn ROV dynamics by combining data-driven predictions with physics-based constraints. 

\subsubsection{One-step-ahead Prediction Loss}
The one-step-ahead prediction loss ($\mathcal{L}_D$) measures the mean squared error between the predicted state $\hat{\mathbf{x}}_{n,m}(T)$ and the ground-truth next state $\mathbf{x}_{n+1,m}(0)$ for each consecutive pair in all trajectories:

\begin{equation}
    \mathcal{L}_D = \frac{1}{N_B(N_D-1)} \sum^{N_B-1}_{m=0} \sum^{N_D-2}_{n=0} ||\mathbf{x}_{n+1,m}(0)-\hat{\mathbf{x}}_{n,m}(T)||_2^2,
\end{equation}
This loss encourages the model to match the known data at discrete intervals $T$.

\subsubsection{Physics Loss}
\begin{figure}[!b]
    \centering
    \includegraphics[width=\columnwidth]{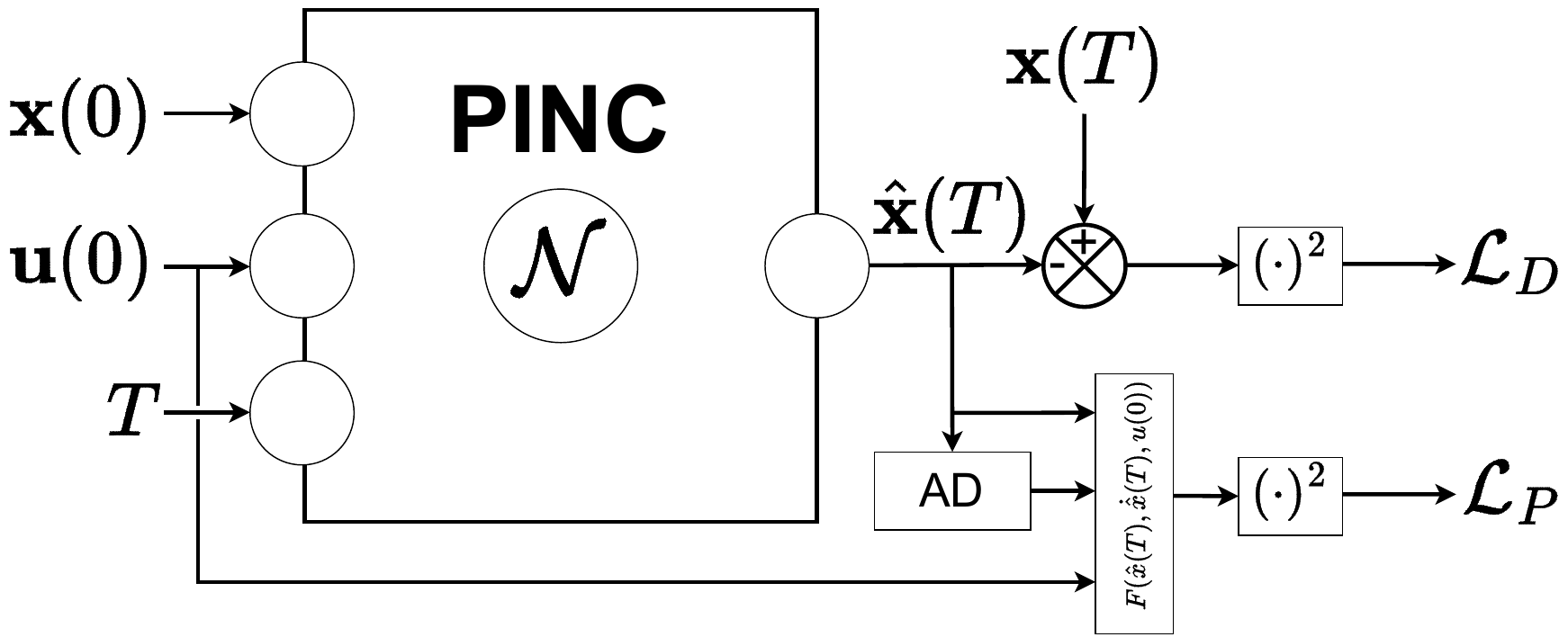}
    \caption{PINC model output used to calculate the MSE data loss by comparing it to the ground truth state and used with AD and the underlying physics function to find the MSE physics loss.}
    \label{fig:pinc_loss_processing}
\end{figure}

The physics loss ($\mathcal{L}_P$) regularizes the model’s predictions to respect the underlying physics by penalizing deviations from the governing differential equations. Fig.~\ref{fig:pinc_loss_processing} illustrates the processing of the data and physics losses within the PINC framework.
Each point in each trajectory has some corresponding collocation points $N_P$, indexed by $k$. Thus, the total number of collocation points evaluated in a batch is $N_D\times N_B\times N_P$. The $k$th collocation point in the $n$th trajectory, at its $m$th point is denoted $T_{n,m,k}^{coll}$, which is sampled by LHS on the interval $T^{coll}\in[0,T]$. The physics residual at each collocation point is defined as:
\begin{equation}
    \label{eq:phy_residual}
    F(\mathbf{x}\dot{,\mathbf{x}},\mathbf{u}) = \dot{\mathbf{x}} - f(\mathbf{x},\mathbf{u}),
\end{equation}
where $f(\mathbf{x},\mathbf{u})$ refers to the right-hand side of the ROV dynamics in~\eqref{eq:rov_simplified}. When $F(\mathbf{x},\dot{\mathbf{x}},\mathbf{u}) =0$, the learned dynamics perfectly match the true dynamics. Specifically the physics residual $F\left(\cdot\right) = F\left(\hat{\mathbf{x}}_{n,m}(T_{n,m,k}^{\text{coll}}), \dot{\hat{\mathbf{x}}}_{n,m}(T_{n,m,k}^{\text{coll}}), \mathbf{u}_{n,m}(0)\right)$ is used in the physics loss. The physics loss is then computed as the mean squared error of these residuals across all collocation points, trajectories, and time steps: 

\begin{equation}
\label{eq:phys_loss}
    \mathcal{L}_P = \frac{1}{N_B N_D N_P} \sum_{k=0}^{N_P - 1} \sum_{m=0}^{N_B - 1} \sum_{n=0}^{N_D - 1} \left\| F\left(\cdot\right) \right\|_2^2.
\end{equation}

\subsubsection{Initial Condition Loss}
The initial condition loss $\mathcal{L}_{IC}$ ensures internal consistency of the network’s outputs at $t=0$, treating each data point as a new initial condition. This is formulated as follows:
\begin{equation}
    \mathcal{L}_{IC} = \frac{1}{N_B N_D} \sum^{N_B-1}_{m=0}\sum^{N_D-1}_{n=0}||\mathbf{x}_{n,m}(0)-\hat{\mathbf{x}}_{n,m}(0)||_2^2.
\end{equation}

\subsubsection{Rollout Loss}
Inspired by \cite{zhao_research_2024}, an $N$-step-ahead (rollout) loss is used to penalize the accumulation of errors when predicting multiple steps forward. In a trajectory with $N_D$, points, only the first $N_R=N_D-N_{pred}$ points can be rolled out. For each trajectory, rollouts are initiated from the first $N_R$ points, and the predicted sequence is compared to the ground truth trajectory using the rollout loss function:
\begin{equation}
    \mathcal{L}_R = \frac{1}{{N_B N_R N_P}} \sum^{N_P}_{k=1} \sum^{N_B-1}_{m=0} \sum^{N_R-1}_{n=0} ||\mathbf{x}_{n+k,m}(0) - \hat{\mathbf{x}}_{n,m}(kT)||_2^2.
\end{equation}

%\subsubsection{Physics Rollout Loss}
%When both the rollout loss and physics loss are included in the optimization, a physics rollout loss is also computed and incorporated to regularize the intermediate states in the multistep predictions, ensuring adherence to the governing differential equations.
\subsubsection{Physics Rollout Loss}
When both the rollout loss and physics loss are included in the optimization, a physics rollout loss is also computed to regularize the intermediate states in multistep predictions. This loss applies the physics residual in~\eqref{eq:phy_residual} to each state in the rollout sequence:
\begin{equation}
    \mathcal{L}_{PR} = \frac{1}{N_{roll}} \sum_{i=0}^{N_{roll}-1} \mathcal{L}_P.
\end{equation}
This term helps ensure that predicted states in the rollout sequence remain consistent with the underlying physical dynamics, preventing error accumulation over multiple steps.

%\textcolor{blue}{where $N_{roll}$ is the number of rollout steps and $\mathcal{L}_P$ is the physics loss calculated at each step as defined in Eq.~\eqref{eq:phys_loss}.}

\subsection{Gradient Weighting}
\label{sec:grad_weighting}
Multitask learning is addressed using the three methods following methods for combining gradients. The first method, simple summation, directly adds the gradients from each loss term. The second approach, ConFIG \cite{liu_config_2024}, dynamically adjusts and reorients gradients so that the combined gradient does not conflict with any individual gradient. It also scales the magnitude based on cosine similarity, making it larger (or smaller) depending on how aligned or misaligned the losses are. The final method, gradient normalization, normalizes each gradient $\mathbf{g}_n$ to match the norm of a “reference” gradient (usually the data-loss gradient $\mathbf{g}_0$), and is then weighed by user-set coefficients $w_n$. Finally, the sum is re-normalized to match the reference gradient’s norm:
    \begin{equation}
    \label{eq:grad_norm_scaling}
        \bar{\mathbf{g}}_n
        =
        \mathbf{g}_n\,
        \frac{\|\mathbf{g}_0\|_2}{\|\mathbf{g}_n\|_2},
        \quad
        \mathbf{g}
        =
        \sum_{n=0}^{N-1} w_n\,\bar{\mathbf{g}}_n,
        \quad
        \bar{\mathbf{g}}
        =
        \mathbf{g}\,
        \frac{\|\mathbf{g}_0\|_2}{\|\mathbf{g}\|_2}.
    \end{equation}
In addition, the norm of the combined gradient is clipped at $||\bar{\mathbf{g}}||_2\leq c_{|g|,max}=5.0$ to avoid huge steps due to high variance or numeric issues.

\subsection{Model Evaluation}
\label{sec:model_eval}
The learned models are evaluated in two ways:
\subsubsection{Accuracy of Predictions}
\begin{itemize}
    \item \textit{One-step-ahead} prediction error on the development set ($\mathcal{L}_{data,dev}$) and test set ($\mathcal{L}_{data,test,interp}$ for interpolation and $\mathcal{L}_{data,test,extrap}$ for extrapolation).
    \item \textit{Rollout} prediction error ($\mathcal{L}_{roll,dev}$) over $10$ steps, capturing cumulative error in sequential predictions.
\end{itemize}
\subsubsection{Long-Horizon Prediction Validity}
The valid prediction time (VPT) is the largest time interval over which the model’s predicted position error remains below a threshold, i.e.

\begin{equation}
\|\mathbf{e}_{\text{pos}}(t)\|_2 = \sqrt{\Delta_x^2 + \Delta_y^2 + \Delta_z^2} \leq 0.05\,\mathrm{m}, 
\end{equation}
where $\Delta_x = x(t) - \hat{x}(t)$, $\Delta_y = y(t) - \hat{y}(t)$ and $\Delta_z = z(t) - \hat{z}(t)$. For all $t < \text{VPT}$, the predictions are considered valid up to that horizon. Once the error exceeds $0.05\,\mathrm{m}$, the prediction is deemed unsafe or unusable for control.

To evaluate the VPT, each trajectory is rolled out from its first initial condition in full, using 
$N_{steps} = 65$, so it matches the ground-truth trajectory length. The VPT is reported in seconds but can readily be converted to discrete steps by dividing by the sampling interval $T$. This metric is computed on the development, interpolation, and extrapolation test sets.

The interpolation and extrapolation test sets are created using time interval sizes $T$ different from those used for training and validation while maintaining the same input types and initial condition intervals as the validation set. To assess interpolation, the time interval size is decreased, testing the model's robustness and its ability to learn the differential equation solution over the entire interval $t\in[0,T]$, aided by the physics loss. For extrapolation, the interval size $T$. is increased. Test sets are generated with the following time interval sizes:
\begin{equation}
\begin{cases}
    T_{test,interp} &= T - 0.25T = 0.08 - 0.02 = 0.06 \\
    T_{test,extrap} &= T + 0.25T = 0.08 + 0.02 = 0.1.
\end{cases}
\end{equation}

%%%%%%%%%%%%%%%%%%%%%%%%%%%%%%%%%%%%%%%%%%%%%%%%%%%%%%%%%%%%%%%%%%%%%%%%%%%%%%%%%%%%%%%%%%%%%%%%%%%%
%%%%%%%%%%%%%%%%%%%%%%%%%%%%%%%%%%%%%%%%%%%%%%%%%%%%%%%%%%%%%%%%%%%%%%%%%%%%%%%%%%%%%%%%%%%%%%%%%%%%
%%%%%%%%%%%%%%%%%%%%%%%%%%%%%%%%%%%%%%%%%%%%%%%%%%%%%%%%%%%%%%%%%%%%%%%%%%%%%%%%%%%%%%%%%%%%%%%%%%%%

\section{Simulation Results} % ~3 pages
\label{sec:results}
A fully translational model with yaw rotation was selected because it incorporates quadratic damping, rotational non-linearities, and gravitational and buoyancy forces. The quadratic non-linearities arise from damping effects. At the same time, the rotational non-linearity in yaw, arising from the planar velocities $u$ and $v$ being rotated from the body frame $\mathcal{F}_B$ to the world frame $\mathcal{F}_W$, in the position dynamics, introduces a limited form of rotational dynamics, allowing to test rotational effects.

%Working with this model aims to explore the viability of using PINC to model underwater ROV dynamics, potentially offering an alternative to analytical first-principles models in model predictive control.

%Although extending the approach to include all rotational degrees of freedom would provide a more comprehensive representation, it necessitates a significant design decision regarding best representing the ROV’s complete rotational dynamics so that the PINC can learn them effectively. This work does not address that aspect. 

\begin{remark}
All quantities are provided in SI units, i.e., time is in seconds [s], and position is in meters [m].
\end{remark}

\begin{remark}
For ease of comparing, losses $\mathcal{L}_D$, $\mathcal{L}_R$, and $\mathcal{L}_P$ are reported on a $\log_{10}(\cdot)$ scale.
\end{remark}

\begin{remark}
The loss terms in the figures are labeled as follows: $\mathcal{L}_{data,dev}$ ($L_1$), $\mathcal{L}_{roll,dev}$ ($L_2$), $\mathcal{L}_{phy,dev}$ ($L_3$), $\mathcal{L}_{data,test,interp}$ ($L_4$), and $\mathcal{L}_{data,test,extrap}$ ($L_5$). The metrics $\text{VPT}_{dev}$, $\text{VPT}_{test,interp}$, and $\text{VPT}_{test,extrap}$ are labeled as $\text{VPT}_1$, $\text{VPT}_2$, and $\text{VPT}_3$, respectively.
\end{remark}

\subsection{Data Generation}
A dataset was generated and partitioned into training, development (validation), and test sets, each with distinct initial conditions and input types. Numerical integration of the dynamics \eqref{eq:rov_simplified} was performed using trajectories starting from randomly sampled initial states $\bm{x}_0$. Each state variable was independently drawn from uniform intervals: position $(x, y, z) \in [-x_{\max}, x_{\max}]$, $[-y_{\max}, y_{\max}]$, $[-z_{\max}, z_{\max}]$; heading angle $\psi \in [-\pi, \pi]$; linear velocities $(u, v, w)$ within specified bounds with $w \geq 0$ to simulate diving behavior; and yaw rate $r$ within a designated range.

Control input sequences $\mathbf{u}(t)$ for the training set were generated using ramp-based patterns with random signs, offsets, and ramp-up/ramp-down profiles. Sine waves with fixed amplitude, random frequencies, and phases were used for the development and test sets. Each input channel was scaled to ensure realistic input magnitudes, as detailed in the respective experiments.

Trajectories were integrated over a total time of $T_{tot}=5.2$ seconds with a sampling period of $T=0.08$ seconds, resulting in $N_{steps}=66$ points per trajectory. To maintain continuity in yaw, $\psi$ was represented by its sine and cosine values, i.e., $(\cos(\psi), \sin(\psi))$ instead of $\psi$ directly. For physics loss computation, additional collocation points $\tau \in [0, T]$ were sampled within each sampling interval using Latin Hypercube Sampling (LHS). The number and placement of collocation points ($N_P$) were adjustable to enforce the governing physics throughout each interval.

\subsection{Experiment Protocol}
A series of experiments is conducted to examine how different design factors influence the performance of PINC. By systematically varying these factors, the aim is to identify PINC configurations that yield the best trade-offs between accuracy, robustness, and computational efficiency. We vary:
\begin{enumerate}

    \item \textbf{Neural Network Size:} 
        Varying the number of layers and neurons to balance expressiveness and complexity.
    
    \item \textbf{Residual Connection Effect:} 
        Assessing the impact of an integral-form residual connection on model accuracy.
    
    \item \textbf{Activation Function Type:} 
        Comparing adaptive activation functions to identify optimal non-linearities.
    
    \item \textbf{Batch Size Variations:}
    Investigating batch size as a regularizer and its effect on convergence speed.
    
    \item \textbf{Physics Loss:} 
        Evaluating whether embedding known physical dynamics in the loss improves model generalization.
    
    \item \textbf{Rollout Loss:}
    Studying how penalizing multi-step prediction affects long-horizon accuracy.
    
    \item \textbf{Initial Condition Loss:}
        Ensuring consistency at $T=0$ to improve predictions at future time steps.
    
    %\item \textbf{Structural Information:}
    %    Rotation of planar position increments to simplify local-body-frame learning.
    
    \item \textbf{Gradient Weighting:}
        Testing summation, ConFIG, and normalization-based schemes for combining gradients.
    
    \item \textbf{Physics Collocation:}
    Adjusting the number and placement of collocation points to enforce system dynamics.
    
    \item \textbf{Noisy Inputs:}
    Injecting Gaussian noise to evaluate the model’s resilience to sensor and measurement errors.
    
    \item \textbf{Learning Rate Scheduling:}
    Exploring if a decaying learning rate improves loss convergence.
    
\end{enumerate}

\subsection{Data and Training Parameters}
\label{subsec:data_training_params}
Unless otherwise stated, the following hyperparameters are used:
\begin{itemize}
    \item Models are trained for $N_{\text{epoch}} = 1200$ epochs (or more if needed for convergence).
    \item A batch size of $N_{\text{batch}} = 3$ is used (unless varied in experiments on batch size).
    \item The AdamW optimizer is applied with an initial learning rate of $\text{lr}_0 = 8\times10^{-3}$ and no scheduling, except when explicitly stated, where the scheduler uses a 'minimum value of $\text{lr}_{\min}=10^{-4}$ with a "patience" of $100$.
    \item Layer normalization is employed on every second layer.
    \item Gradient weighting is applied using a normalization scheme with weights: $\{w_{\text{data}} = 1.0, w_{\text{roll}} = 1.0, w_{\text{phy}} = 0.5, w_{\text{phy,roll}} = 0.5, w_{\text{ic}} = 0.5\}$.
    \item The output planar position change is rotated by the predicted yaw (unless the structural information is ablated).
\end{itemize}

The models are trained using a dataset of $N_{\text{traj,data}} = 400$ trajectories with ramp-based inputs. The development set initially contains $N_{\text{traj,dev}} = 80$ trajectories, but for evaluation, is increased to $N_{\text{traj,dev}} = 1000$ to match the size of the interpolation and extrapolation test sets. Both the development and test sets use sinusoidal inputs. The number of steps is kept at $N_{\text{steps}} = 66$, and the total time in the test sets is varied to ensure consistent trajectory lengths.

For all experiments except the last two, the training data are generated under the following initial condition ranges: $x_{\max} = y_{\max} = z_{\max} = 1.0$, $\psi_{\max} = \pi$, $u_{\max} = 1.0$, $v_{\max} =0.0$, $w_{\max} = 0.1$, $r_{\max} = 0.0$. The inputs for training are ramp-based rather than step-based. Before any sign, scaling, or offset is applied, each ramp starts at $0$, increases to $1$ (peaking at $0.5T$), and then decreases back to $0$. The offsets are sampled from a zero mean normal distribution with $\sigma^2=0.25$, the ramp's sign from a Bernoulli random variable with $p=0.5$. 
All experiments are conducted using a single seed for randomness in data generation and network initialization. The seed is fixed at 0 for reproducibility.

The development and test sets are generated with $x_{\max} = y_{\max} = z_{\max} = u_{\max} = v_{\max} = w_{\max} = r_{\max} = 0.0$ and $\psi_{\max} = \pi$, so that the models are evaluated only on their ability to handle different inputs rather than different initial conditions. The inputs are sinusoids with an amplitude of $A=3$, a random frequency sampled uniformly in the interval $[0.01, 0.2]$, and a random phase uniformly sampled in the interval $[0, 2\pi]$.

\begin{remark}
The development set used in all experiments is generated independently, and it is uncorrelated with the training data.
\end{remark}

\begin{remark}
    The inputs in all datasets are scaled as follows: $Y$ is scaled down by $0.1$, $M_z$ is scaled down by $0.05$, and $Z$ is scaled up by $5$ and then taken as an absolute, ensuring only downward motion is commanded since the ROV's static buoyancy force naturally surfaces the ROV.
\end{remark}

\subsection{Neural Network Architecture Experiments}
\label{subsec:nn_arch}

\newcommand{\PlotHeight}{6cm}

%%%%%%%%%%%%%%%%
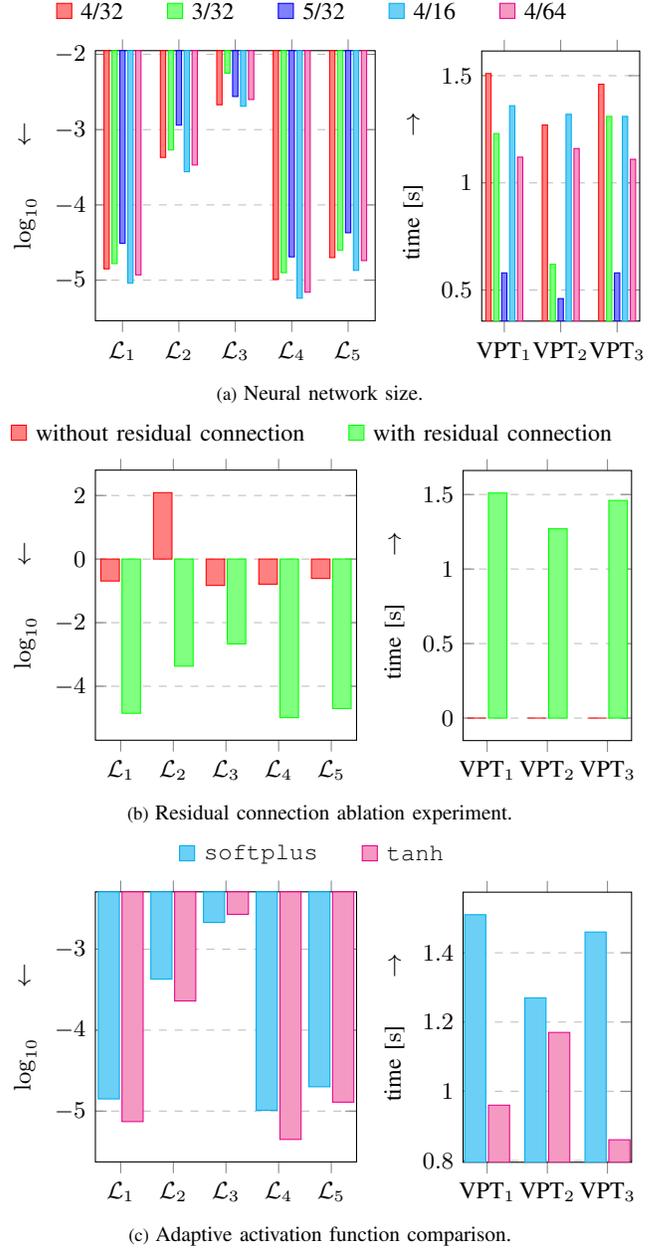
\begin{figure}[!t]
    %%%%%%%%%%%%%%%%%%%%%%%%%%%%%%%%%%%%%%%%%%%%%%%%%%
%%%%%%%%%%%%%%%%%%%%%%%%%%%%%%%%%%%%%%%%%%%%%%%%%%
% Neural network size experiment
%%%%%%%%%%%%%%%%%%%%%%%%%%%%%%%%%%%%%%%%%%%%%%%%%
%%%%%%%%%%%%%%%%%%%%%%%%%%%%%%%%%%%%%%%%%%%%%%%%%

{\pgfplotstableread[col sep=comma]{%
    Result, {4/32}, {3/32}, {5/32}, {4/16}, {4/64}
    $\mathcal{L}_1$, -4.85, -4.78, -4.51, -5.04, -4.93
    $\mathcal{L}_2$, -3.37, -3.27, -2.94, -3.56, -3.47
    $\mathcal{L}_3$, -2.67, -2.25, -2.56, -2.69, -2.60
    $\mathcal{L}_4$, -4.99, -4.90, -4.69, -5.24, -5.16
    $\mathcal{L}_5$, -4.70, -4.60, -4.37, -4.87, -4.74
}\tableL

\pgfplotstableread[col sep=comma]{%
    Result, {4/32}, {3/32}, {5/32}, {4/16}, {4/64}
    $\text{VPT}_1$, 1.51, 1.23, 0.58, 1.36, 1.12
    $\text{VPT}_2$, 1.27, 0.62, 0.46, 1.32, 1.16
    $\text{VPT}_3$, 1.46, 1.31, 0.58, 1.31, 1.11
}\tableVPT

\centering
\begin{tikzpicture}
    \matrix[matrix of nodes, nodes={align=left, anchor=west, font=\small}]
        {
            \fill[draw=red,fill=red!50] (0,-0.1) rectangle (0.2,0.1); & {4/32} & [10pt]
            \fill[draw=green,fill=green!50] (0,-0.1) rectangle (0.2,0.1); & {3/32} & [10pt]
            \fill[draw=blue,fill=blue!50] (0,-0.1) rectangle (0.2,0.1); & {5/32} & [10pt]
            \fill[draw=cyan,fill=cyan!50] (0,-0.1) rectangle (0.2,0.1); & {4/16} & [10pt]
            \fill[draw=magenta,fill=magenta!50] (0,-0.1) rectangle (0.2,0.1); & {4/64} \\
        };
\end{tikzpicture}

%%%%%%%%%%%%%%%%%%%%%%%%%%%%%%%%%%%%%%%%%%%
\setcounter{subfigure}{0}
\subfloat{
    \subfloat{
        \begin{tikzpicture}
    \begin{axis}[
        every axis plot post/.style={/pgf/number format/fixed},
        width=\PlotHeight,
        ybar=1,
        x=0.75cm,
        ylabel near ticks,
        xtick=data,
        enlarge x limits=0.12,
        bar width=2,
        ymajorgrids=true,
        grid style=dashed,
        xticklabel style={rotate=0},
        label style={font=\small},
        tick label style={font=\small},
        ylabel={$\log_{10}~\quad~\leftarrow$},
        symbolic x coords={$\mathcal{L}_1$,$\mathcal{L}_2$,$\mathcal{L}_3$,$\mathcal{L}_4$,$\mathcal{L}_5$}
        ]
    \addplot[draw=red,fill=red!50] table[x=Result, y={4/32}] {\tableL};
    \addplot[draw=green,fill=green!50] table[x=Result, y={3/32}] {\tableL};
    \addplot[draw=blue,fill=blue!50] table[x=Result, y={5/32}] {\tableL};
    \addplot[draw=cyan,fill=cyan!50] table[x=Result, y={4/16}] {\tableL};
    \addplot[draw=magenta,fill=magenta!50] table[x=Result, y={4/64}] {\tableL};
    \end{axis}
        \end{tikzpicture}
        \label{fig:b2d_center}}
    \hfill
    \subfloat{
        \begin{tikzpicture}
    \begin{axis}[
        every axis plot post/.style={/pgf/number format/fixed},
        width=\PlotHeight,
        ybar=1,
        x=0.75cm,
        ylabel near ticks,
        xtick=data,
        enlarge x limits=0.2,
        bar width=2,
        ymajorgrids=true,
        grid style=dashed,
        xticklabel style={rotate=0},
        label style={font=\small},
        tick label style={font=\small},
        ylabel={time [\si{s}]$~\quad~\rightarrow$},
        symbolic x coords={$\text{VPT}_1$,$\text{VPT}_2$,$\text{VPT}_3$}
        ]
    \addplot[draw=red,fill=red!50] table[x=Result, y={4/32}] {\tableVPT};
    \addplot[draw=green,fill=green!50] table[x=Result, y={3/32}] {\tableVPT};
    \addplot[draw=blue,fill=blue!50] table[x=Result, y={5/32}] {\tableVPT};
    \addplot[draw=cyan,fill=cyan!50] table[x=Result, y={4/16}] {\tableVPT};
    \addplot[draw=magenta,fill=magenta!50] table[x=Result, y={4/64}] {\tableVPT};
    \end{axis}
    %\subcaption{Neural network size}
        \end{tikzpicture}
 %       \setcounter{subfigure}{1}
%\subcaption{Neural network size}
        \label{fig:neural_size}}
}
\setcounter{subfigure}{0}

\subcaption{\footnotesize{Neural network size.}}

%%%%%%%%%%%%%%%%%%%%%%%%%%%%%%%%%%%%%%%%%%%%%%%%%%
%%%%%%%%%%%%%%%%%%%%%%%%%%%%%%%%%%%%%%%%%%%%%%%%%%
% Residual connection ablation experiment
%%%%%%%%%%%%%%%%%%%%%%%%%%%%%%%%%%%%%%%%%%%%%%%%%
%%%%%%%%%%%%%%%%%%%%%%%%%%%%%%%%%%%%%%%%%%%%%%%%%

%\vspace{0.5cm}

\pgfplotstableread[col sep=comma]{%
    Result, {with residual connection}, {without residual connection}
    $\mathcal{L}_1$, -4.85, -0.69
    $\mathcal{L}_2$, -3.37, 2.09
    $\mathcal{L}_3$, -2.67, -0.83
    $\mathcal{L}_4$, -4.99, -0.79 
    $\mathcal{L}_5$, -4.70, -0.61
}\tableL

\pgfplotstableread[col sep=comma]{%
    Result, {with residual connection}, {without residual connection}
    $\text{VPT}_1$, 1.51, 0.0
    $\text{VPT}_2$, 1.27, 0.0
    $\text{VPT}_3$, 1.46, 0.0
}\tableVPT

\centering
%\begin{center}
\begin{tikzpicture}
    \matrix[matrix of nodes, nodes={align=left, anchor=west, font=\small}]
        {
            \fill[draw=red,fill=red!50] (0,-0.1) rectangle (0.2,0.1); & {without residual connection} & [10pt]
            \fill[draw=green, fill=green!50] (0,-0.1) rectangle (0.2,0.1); & {with residual connection} \\
        };
\end{tikzpicture}
%\end{center}

%% Log Error
\subfloat{
    \subfloat{
        \begin{tikzpicture}
    \begin{axis}[
        every axis plot post/.style={/pgf/number format/fixed},
        width=\PlotHeight,
        ybar=1,
        x=0.7cm,
        ylabel near ticks,
        xtick=data,
        enlarge x limits=0.12,
        bar width=7,
        ymajorgrids=true,
        grid style=dashed,
        xticklabel style={rotate=0},
        label style={font=\small},
        tick label style={font=\small},
        ylabel={$\log_{10}~\quad~\leftarrow$},
        symbolic x coords={$\mathcal{L}_1$,$\mathcal{L}_2$,$\mathcal{L}_3$,$\mathcal{L}_4$,$\mathcal{L}_5$}
        ]
    \addplot[draw=red,fill=red!50] table[x=Result, y={without residual connection}] {\tableL};
    \addplot[draw=green,fill=green!50] table[x=Result, y={with residual connection}]  {\tableL};
    \end{axis}
        \end{tikzpicture}
        \label{fig:logE}}
    \hfill
%% VPT
    \subfloat{
        \begin{tikzpicture}
    \begin{axis}[
        every axis plot post/.style={/pgf/number format/fixed},
        width=\PlotHeight,
        ybar=1,
        x=0.8cm,
        ylabel near ticks,
        xtick=data,
        enlarge x limits=0.2,
        bar width=7,
        ymajorgrids=true,
        grid style=dashed,
        xticklabel style={rotate=0},
        label style={font=\small},
        tick label style={font=\small},
        ylabel={time [\si{s}]$~\quad~\rightarrow$},
        symbolic x coords={$\text{VPT}_1$,$\text{VPT}_2$,$\text{VPT}_3$}
        ]
    \addplot[draw=red,fill=red!50] table[x=Result, y={without residual connection}] {\tableVPT};
    \addplot[draw=green,fill=green!50] table[x=Result, y={with residual connection}]  {\tableVPT};
    \end{axis}
        \end{tikzpicture}
        \label{fig:neural_s}}   
}

\setcounter{subfigure}{1}

\subcaption{\footnotesize{Residual connection ablation experiment.}}

%\subcaption{Residual connection ablation experiment}

%%%%%%%%%%%%%%%%%%%%%%%%%%%%%%%%%%%%%%%%%%%%%%%%%%
%%%%%%%%%%%%%%%%%%%%%%%%%%%%%%%%%%%%%%%%%%%%%%%%%%
%  Adaptive activation function comparison
%%%%%%%%%%%%%%%%%%%%%%%%%%%%%%%%%%%%%%%%%%%%%%%%%
%%%%%%%%%%%%%%%%%%%%%%%%%%%%%%%%%%%%%%%%%%%%%%%%%

\pgfplotstableread[col sep=comma]{%
    Result, {with residual connection}, {without residual connection}
    $\mathcal{L}_1$, -4.85, -5.13
    $\mathcal{L}_2$, -3.37, -3.64
    $\mathcal{L}_3$, -2.67, -2.57
    $\mathcal{L}_4$, -4.99, -5.35 
    $\mathcal{L}_5$, -4.70, -4.89
}\tableL

\pgfplotstableread[col sep=comma]{%
    Result, {with residual connection}, {without residual connection}
    $\text{VPT}_1$, 1.51, 0.96
    $\text{VPT}_2$, 1.27, 1.17
    $\text{VPT}_3$, 1.46, 0.86
}\tableVPT

\centering
%\begin{center}
\begin{tikzpicture}
    \matrix[matrix of nodes, nodes={align=left, anchor=west, font=\small}]
        {
            \fill[draw=cyan, fill=cyan!50] (0,-0.1) rectangle (0.2,0.1); & \lstinline|softplus| & [10pt]
            \fill[draw=magenta, fill=magenta!50] (0,-0.1) rectangle (0.2,0.1); & \lstinline|tanh| \\
        };
\end{tikzpicture}
%\end{center}

%% Log Error
\subfloat{
    \subfloat{
        \begin{tikzpicture}
    \begin{axis}[
        every axis plot post/.style={/pgf/number format/fixed},
        width=\PlotHeight,
        ybar=1,
        x=0.7cm,
        ylabel near ticks,
        xtick=data,
        enlarge x limits=0.12,
        bar width=8,
        ymajorgrids=true,
        grid style=dashed,
        xticklabel style={rotate=0},
        label style={font=\small},
        tick label style={font=\small},
        ylabel={$\log_{10}~\quad~\leftarrow$},
        symbolic x coords={$\mathcal{L}_1$,$\mathcal{L}_2$,$\mathcal{L}_3$,$\mathcal{L}_4$,$\mathcal{L}_5$}
        ]
    \addplot[draw=cyan,fill=cyan!50] table[x=Result, y={with residual connection}] {\tableL};
    \addplot[draw=magenta,fill=magenta!50] table[x=Result, y={without residual connection}]  {\tableL};
    \end{axis}
        \end{tikzpicture}
        \label{fig:logE}}
    \hfill
%% VPT
    \subfloat{
        \begin{tikzpicture}
    \begin{axis}[
        every axis plot post/.style={/pgf/number format/fixed},
        width=\PlotHeight,
        ybar=1,
        x=0.8cm,
        ylabel near ticks,
        xtick=data,
        enlarge x limits=0.2,
        bar width=8,
        ymajorgrids=true,
        grid style=dashed,
        xticklabel style={rotate=0},
        label style={font=\small},
        tick label style={font=\small},
        ylabel={time [\si{s}]$~\quad~\rightarrow$},
        symbolic x coords={$\text{VPT}_1$,$\text{VPT}_2$,$\text{VPT}_3$}
        ]
    \addplot[draw=cyan,fill=cyan!50] table[x=Result, y={with residual connection}] {\tableVPT};
    \addplot[draw=magenta,fill=magenta!50] table[x=Result, y={without residual connection}]  {\tableVPT};
    \end{axis}
        \end{tikzpicture}
       \label{fig:residual}}      
}%
\setcounter{subfigure}{2}%
\label{fig:adptation}%
\subcaption{\footnotesize{Adaptive activation function comparison.}}%
}%
    \caption{Overview of architecture experimental results: (a) Neural Network Size: $N_L = 4$ and $N_H = 32$ yield the lowest losses and highest VPTs. (b)~Residual Connection: Ablating the residual connection causes a significant performance drop, with $\mathcal{L}_{data,dev}$ over four orders of magnitude higher. (c)~Activation Function: \lstinline|softplus| achieves better VPTs than \lstinline|tanh|.}
    \label{fig:architecture}
\end{figure}
%%%%%%%%%%%%%%%%

\subsubsection{Neural Network Size}
The number of hidden layers $N_L$ and the number of neurons $N_H$ in each layer is varied in this experiment, which performance metrics are shown in Fig.~\ref{fig:neural_s}. It is found that all losses are the lowest and VPTs the highest for the configuration with $N_L = 4$ and $N_H = 32$. This configuration will, therefore, be used in all subsequent experiments.

\subsubsection{Residual Connection}
The residual connection is ablated in the experiment shown in Fig. \ref{fig:residual}, comparing the model with and without the input state residual connection. Without the residual connection that makes the model follow the integration form of an ODE solution, its performance is degraded, having a $\mathcal{L}_{data,dev}$ that is more than $4$ orders of magnitude higher. 
\subsubsection{Activation Function}
The use of an adaptive \lstinline|softplus| activation function is compared with an adaptive \lstinline|tanh| as the model's non-linearity. The hypothesis is that \lstinline|tanh| might yield smaller errors due to its bounded nature but could suffer from vanishing gradients, while \lstinline|softplus| is unbounded and generally avoids those problems.
The results shown in Fig.~\ref{fig:adptation} indicate that while the losses are mostly lower with \lstinline|tanh|, the VPTs are significantly better when using \lstinline|softplus|. A likely explanation is that \lstinline|softplus| remains smooth and avoids saturation.

\subsection{Effect of adding Rollout Loss on Learning}
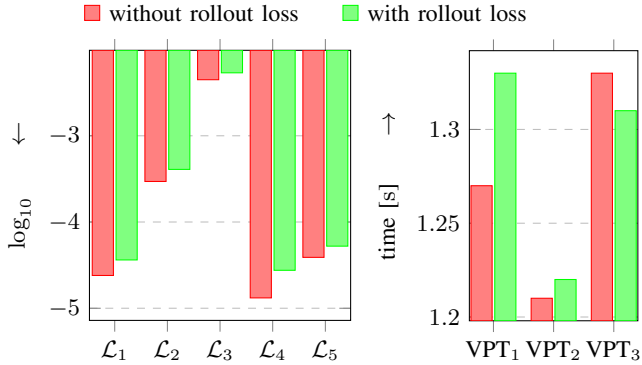
\begin{figure}[!t]
\pgfplotstableread[col sep=comma]{%
    Result, {with residual connection}, {without residual connection}
    $\mathcal{L}_1$, -4.62, -4.44
    $\mathcal{L}_2$, -3.53, -3.39
    $\mathcal{L}_3$, -2.35, -2.27
    $\mathcal{L}_4$, -4.88, -4.56 
    $\mathcal{L}_5$, -4.41, -4.28
}\tableL

\pgfplotstableread[col sep=comma]{%
    Result, {with residual connection}, {without residual connection}
    $\text{VPT}_1$, 1.27, 1.33
    $\text{VPT}_2$, 1.21, 1.22
    $\text{VPT}_3$, 1.33, 1.31
}\tableVPT

\centering
%\begin{center}
\begin{tikzpicture}
    \matrix[matrix of nodes, nodes={align=left, anchor=west, font=\small}]
        {
            \fill[draw=red, fill=red!50] (0,-0.1) rectangle (0.2,0.1); & {without rollout loss} & [10pt]
            \fill[draw=green, fill=green!50] (0,-0.1) rectangle (0.2,0.1); & {with rollout loss} \\
        };
\end{tikzpicture}
%\end{center}

%% Log Error
\subfloat{
    \subfloat{
        \begin{tikzpicture}
    \begin{axis}[
        every axis plot post/.style={/pgf/number format/fixed},
        width=\PlotHeight,
        ybar=1,
        x=0.7cm,
        ylabel near ticks,
        xtick=data,
        enlarge x limits=0.12,
        bar width=8,
        ymajorgrids=true,
        grid style=dashed,
        xticklabel style={rotate=0},
        label style={font=\small},
        tick label style={font=\small},
        ylabel={$\log_{10}~\quad~\leftarrow$},
        symbolic x coords={$\mathcal{L}_1$,$\mathcal{L}_2$,$\mathcal{L}_3$,$\mathcal{L}_4$,$\mathcal{L}_5$}
        ]
    \addplot[draw=red,fill=red!50] table[x=Result, y={with residual connection}] {\tableL};
    \addplot[draw=green,fill=green!50] table[x=Result, y={without residual connection}]  {\tableL};
    \end{axis}
        \end{tikzpicture}
        \label{fig:logE_noise}}
    \hfill
%% VPT
    \subfloat{
        \begin{tikzpicture}
    \begin{axis}[
        every axis plot post/.style={/pgf/number format/fixed},
        width=\PlotHeight,
        ybar=1,
        x=0.8cm,
        ylabel near ticks,
        xtick=data,
        enlarge x limits=0.2,
        bar width=8,
        ymajorgrids=true,
        grid style=dashed,
        xticklabel style={rotate=0},
        label style={font=\small},
        tick label style={font=\small},
        ylabel={time [\si{s}]$~\quad~\rightarrow$},
        symbolic x coords={$\text{VPT}_1$,$\text{VPT}_2$,$\text{VPT}_3$}
        ]
    \addplot[draw=red,fill=red!50] table[x=Result, y={with residual connection}] {\tableVPT};
    \addplot[draw=green,fill=green!50] table[x=Result, y={without residual connection}]  {\tableVPT};
    \end{axis}
        \end{tikzpicture}
       \label{fig:vpt_noise}}      
}
    \caption{Effect of adding rollout loss vs. physics loss on VPT: Incorporating rollout loss with physics loss does not yield any additional performance gains compared to using physics loss alone, as shown in the Loss and VPT metrics.}
    \label{fig:rollout}
\end{figure}

The impact of adding the rollout loss $\mathcal{L}_P$ (Fig. \ref{fig:rollout}) to the overall learning process is explored by comparing the performance of the PINC with and without the rollout loss. While the introduction of the physics loss $\mathcal{L}_P$ alone leads to improvements in several performance metrics, the inclusion of the rollout loss does not consistently enhance the learning. In fact, in some instances, the rollout loss negatively impacts performance, suggesting that the additional regularization from the rollout transformation may not always contribute positively to the overall model training. These results highlight the complex interaction between structural information and physics-based regularization, with the rollout loss sometimes leading to suboptimal learning outcomes.

%%%%%%%%%%%%%%%%

\subsection{Input Noise Robustness Experiment}
\label{subsec:input_noise}
Real-world ROV data often contain sensor or measurement noise. The model’s robustness to noisy inputs is assessed by injecting Gaussian noise, $\bm{\epsilon}$, with a standard deviation of $\sigma = 0.05$ into the neural network's inputs during training.
In each epoch, a noise vector for each state in each trajectory is sampled and injected:
\begin{equation}
    \mathbf{x}_{\text{noisy}} = \mathbf{x} + \bm{\epsilon}.
\end{equation}
In Fig. \ref{fig:noise}, the no-physics and physics+rollout models degrade when noise is introduced. Nonetheless, the physics+rollout model remains more robust, implying that physics regularization can partially compensate for noisy input signals when training the model. The models are evaluated without noise.
% % Please add the following required packages to your document preamble:
% % \usepackage{graphicx}
% \begin{table}[b]
% \centering
% \caption{Training input noise experiment}
% \label{tab:input_noise}
% \resizebox{\columnwidth}{!}{%
% \begin{tabular}{l|r|r|r|r}
% \textbf{Physics and rollout loss included} & \textbf{no} & \textbf{no} & \textbf{yes} & \textbf{yes} \\ \hline
% \textbf{Noise level ($\sigma$)}             & $0.00$        & $0.05$        & $0.00$        & $0.05$        \\ \hline
% $\log_{10}(\mathcal{L}_{data,dev})$         & $-4.45$       & $-3.09$       & $\bm{-4.68}$  & $-3.28$       \\ \hline
% $\log_{10}(\mathcal{L}_{roll,dev})$         & $-2.89$       & $-1.62$       & $\bm{-3.12}$  & $-1.54$       \\ \hline
% $\log_{10}(\mathcal{L}_{phy,dev})$          & $-1.84$       & $-0.64$       & $\bm{-2.69}$  & $-0.83$       \\ \hline
% $\log_{10}(\mathcal{L}_{data,test,interp})$ & $-4.81$       & $-3.03$       & $\bm{-4.88}$  & $-3.24$       \\ \hline
% $\log_{10}(\mathcal{L}_{data,test,extrap})$ & $-4.16$       & $-3.12$       & $\bm{-4.53}$  & $-3.31$       \\ \hline
% $\text{VPT}_{dev}$ [s]/[steps]              & $0.46$/$5.75$ & $0.12$/$1.50$ & $\bm{0.52}$/$6.50$ & $0.20$/$2.50$ \\ \hline
% $\text{VPT}_{test,interp}$ [s]              & $0.42$/$7.00$ & $0.09$/$1.50$ & $\bm{0.46}$/$7.67$ & $0.15$/$2.50$ \\ \hline
% $\text{VPT}_{test,extrap}$ [s]              & $0.48$/$4.80$ & $0.15$/$1.50$ & $\bm{0.54}$/$5.40$ & $0.28$/$2.80$
% \end{tabular}
% }
% \end{table}

%%%%%%%%%%%%%%%%
\begin{figure}[!t]
    \centering
    \pgfplotstableread[col sep=comma]{%
    Result, {with residual connection}, {without residual connection}
    $\mathcal{L}_1$, -3.09, -3.28
    $\mathcal{L}_2$, -1.62, -1.54
    $\mathcal{L}_3$, -0.64, -0.83
    $\mathcal{L}_4$, -3.03, -3.24 
    $\mathcal{L}_5$, -3.12, -3.31
}\tableL

\pgfplotstableread[col sep=comma]{%
    Result, {with residual connection}, {without residual connection}
    $\text{VPT}_1$, 0.12, 0.2
    $\text{VPT}_2$, 0.09, 0.15
    $\text{VPT}_3$, 0.15, 0.28
}\tableVPT

\centering
%\begin{center}
\begin{tikzpicture}
    \matrix[matrix of nodes, nodes={align=left, anchor=west, font=\small}]
        {
            \fill[draw=red, fill=red!50] (0,-0.1) rectangle (0.2,0.1); & {no physics information} & [10pt]
            \fill[draw=green, fill=green!50] (0,-0.1) rectangle (0.2,0.1); & {physics and rollout loss included} \\
        };
\end{tikzpicture}
%\end{center}

%% Log Error
\subfloat{
    \subfloat{
        \begin{tikzpicture}
    \begin{axis}[
        every axis plot post/.style={/pgf/number format/fixed},
        width=\PlotHeight,
        ybar=1,
        x=0.7cm,
        ylabel near ticks,
        xtick=data,
        enlarge x limits=0.12,
        bar width=8,
        ymajorgrids=true,
        grid style=dashed,
        xticklabel style={rotate=0},
        label style={font=\small},
        tick label style={font=\small},
        ylabel={$\log_{10}~\quad~\leftarrow$},
        symbolic x coords={$\mathcal{L}_1$,$\mathcal{L}_2$,$\mathcal{L}_3$,$\mathcal{L}_4$,$\mathcal{L}_5$}
        ]
    \addplot[draw=red,fill=red!50] table[x=Result, y={with residual connection}] {\tableL};
    \addplot[draw=green,fill=green!50] table[x=Result, y={without residual connection}]  {\tableL};
    \end{axis}
        \end{tikzpicture}
        \label{fig:logE_noise}}
    \hfill
%% VPT
    \subfloat{
        \begin{tikzpicture}
    \begin{axis}[
        every axis plot post/.style={/pgf/number format/fixed},
        width=\PlotHeight,
        ybar=1,
        x=0.8cm,
        ylabel near ticks,
        xtick=data,
        enlarge x limits=0.2,
        bar width=8,
        ymajorgrids=true,
        grid style=dashed,
        xticklabel style={rotate=0},
        label style={font=\small},
        tick label style={font=\small},
        ylabel={time [\si{s}]$~\quad~\rightarrow$},
        symbolic x coords={$\text{VPT}_1$,$\text{VPT}_2$,$\text{VPT}_3$}
        ]
    \addplot[draw=red,fill=red!50] table[x=Result, y={with residual connection}] {\tableVPT};
    \addplot[draw=green,fill=green!50] table[x=Result, y={without residual connection}]  {\tableVPT};
    \end{axis}
        \end{tikzpicture}
       \label{fig:vpt_noise}}      
}
    \caption{Effect of noise on learning: Noise was introduced into the ROV simulation model to evaluate the learning performance. Incorporating physics information significantly reduces underfitting and enhances robustness to noise, enabling more reliable learning.}
    \label{fig:noise}
\end{figure}
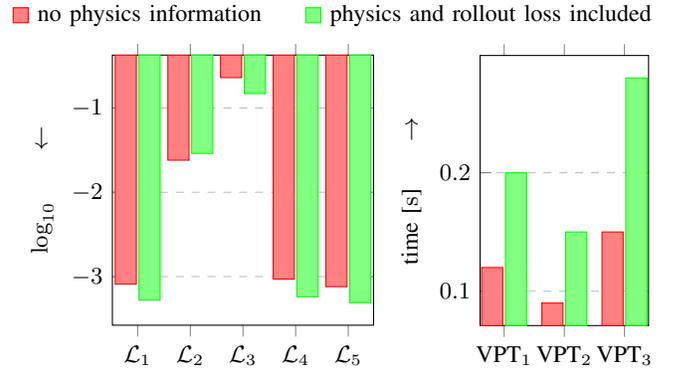
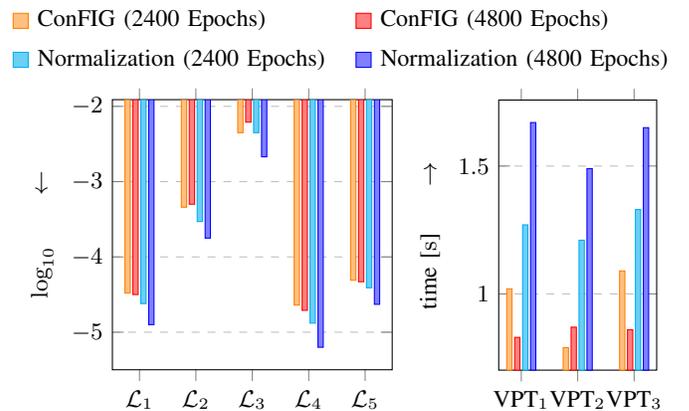
\begin{figure}[!t]
%\vspace{-0.5cm}

%%%%%%%%%%%%%%%%%%%%%%%%%%%%%%%%%%%%%%%%%%%%%%%%%%
%%%%%%%%%%%%%%%%%%%%%%%%%%%%%%%%%%%%%%%%%%%%%%%%%%
% Neural network size experiment
%%%%%%%%%%%%%%%%%%%%%%%%%%%%%%%%%%%%%%%%%%%%%%%%%
%%%%%%%%%%%%%%%%%%%%%%%%%%%%%%%%%%%%%%%%%%%%%%%%%

\pgfplotstableread[col sep=comma]{%
    Result, {ConFig 1}, {ConFig 2}, {Normalization 1}, {Normalization 2}
    $\mathcal{L}_1$, -4.48, -4.50, -4.62, -4.9
    $\mathcal{L}_2$, -3.34, -3.30, -3.53, -3.75
    $\mathcal{L}_3$, -2.35, -2.21, -2.35, -2.67
    $\mathcal{L}_4$, -4.64, -4.71, -4.88, -5.20
    $\mathcal{L}_5$, -4.31, -4.33, -4.41, -4.63
}\tableL

\pgfplotstableread[col sep=comma]{%
    Result, {ConFig 1}, {ConFig 2}, {Normalization 1}, {Normalization 2}
    $\text{VPT}_1$, 1.02, 0.83, 1.27, 1.67
    $\text{VPT}_2$, 0.79, 0.87, 1.21, 1.49
    $\text{VPT}_3$, 1.09, 0.86, 1.33, 1.65
}\tableVPT

\centering

\begin{tikzpicture}
\matrix[matrix of nodes, nodes={align=left, anchor=west, font=\small}] {
    \fill[draw=orange,fill=orange!50] (0,-0.1) rectangle (0.2,0.1); & ConFIG (2400 Epochs) & [5pt]
    \fill[draw=red,fill=red!50] (0,-0.1) rectangle (0.2,0.1); & ConFIG (4800 Epochs) \\
    \fill[draw=cyan,fill=cyan!50] (0,-0.1) rectangle (0.2,0.1); & Normalization (2400 Epochs) & [5pt]
    \fill[draw=blue,fill=blue!50] (0,-0.1) rectangle (0.2,0.1); & Normalization (4800 Epochs) \\
};
\end{tikzpicture}

%%%%%%%%%%%%%%%%%%%%%%%%%%%%%%%%%%%%%%%%%%%

\subfloat{
    \subfloat{
        \begin{tikzpicture}
    \begin{axis}[
        every axis plot post/.style={/pgf/number format/fixed},
        width=\PlotHeight,
        ybar=1,
        x=0.75cm,
        ylabel near ticks,
        xtick=data,
        enlarge x limits=0.12,
        bar width=2,
        ymajorgrids=true,
        grid style=dashed,
        xticklabel style={rotate=0},
        label style={font=\small},
        tick label style={font=\small},
        ylabel={$\log_{10}~\quad~\leftarrow$},
        symbolic x coords={$\mathcal{L}_1$,$\mathcal{L}_2$,$\mathcal{L}_3$,$\mathcal{L}_4$,$\mathcal{L}_5$}
        ]
    \addplot[draw=orange,fill=orange!50] table[x=Result, y={ConFig 1}] {\tableL};
    \addplot[draw=red,fill=red!50] table[x=Result, y={ConFig 2}] {\tableL};
    \addplot[draw=cyan,fill=cyan!50] table[x=Result, y={Normalization 1}] {\tableL};
    \addplot[draw=blue,fill=blue!50] table[x=Result, y={Normalization 2}] {\tableL};
    \end{axis}
        \end{tikzpicture}
        \label{fig:b2d_center}}
    \hfill
    \subfloat{
        \begin{tikzpicture}
    \begin{axis}[
        every axis plot post/.style={/pgf/number format/fixed},
        width=\PlotHeight,
        ybar=1,
        x=0.75cm,
        ylabel near ticks,
        xtick=data,
        enlarge x limits=0.2,
        bar width=2,
        ymajorgrids=true,
        grid style=dashed,
        xticklabel style={rotate=0},
        label style={font=\small},
        tick label style={font=\small},
        ylabel={time [\si{s}]$~\quad~\rightarrow$},
        symbolic x coords={$\text{VPT}_1$,$\text{VPT}_2$,$\text{VPT}_3$}
        ]
    \addplot[draw=orange,fill=orange!50] table[x=Result, y={ConFig 1}] {\tableVPT};
    \addplot[draw=red,fill=red!50] table[x=Result, y={ConFig 2}] {\tableVPT};
    \addplot[draw=cyan,fill=cyan!50] table[x=Result, y={Normalization 1}] {\tableVPT};
    \addplot[draw=blue,fill=blue!50] table[x=Result, y={Normalization 2}] {\tableVPT};
    \end{axis}
        \end{tikzpicture}
        %\label{fig:nnsize}    
\label{fig:nnsize}}    
}
\setcounter{subfigure}{0}
%\subcaption{Neural Network size}
%%%%%%%%%%%%%%%%%%%%%%%%%%%%%%%%%%%%%%%%%%%%%%%%%%
%%%%%%%%%%%%%%%%%%%%%%%%%%%%%%%%%%%%%%%%%%%%%%%%%%
% Residual connection ablation experiment
%%%%%%%%%%%%%%%%%%%%%%%%%%%%%%%%%%%%%%%%%%%%%%%%%
%%%%%%%%%%%%%%%%%%%%%%%%%%%%%%%%%%%%%%%%%%%%%%%%%
    \caption{Comparison of PINC models trained with ConFIG and gradient normalization techniques for 2400 and 4800 epochs, using the best network architecture configuration determined from prior ablation studies. Results show that our proposed normalization method consistently outperforms the state-of-the-art ConFIG method in terms of VPT and loss reduction.}
    \label{fig:final_results}
\end{figure}
%%%%%%%%%%%%%%%%

\definecolor{myred}{rgb}{1.0, 0.1, 0.1}  % Pinkish red
\definecolor{mygreen}{rgb}{0.1, 1, 0.1}  % Light green
\definecolor{myblue}{rgb}{0.1, 0.1, 1.0}    % Cyan-ish blue

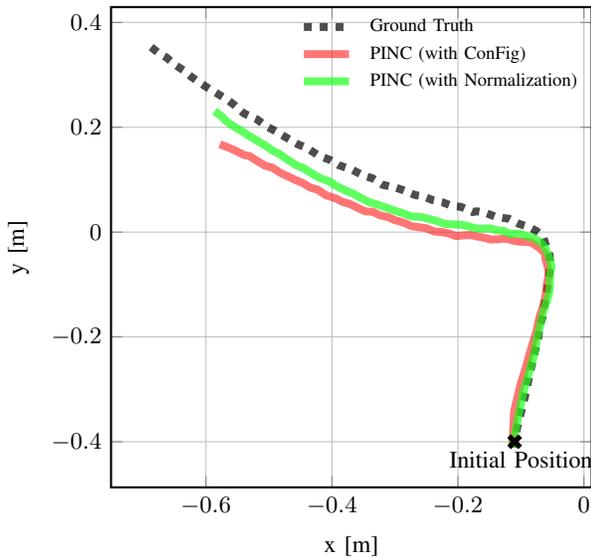
\begin{figure}[!t]
    \centering
    \begin{tikzpicture}
    \begin{axis}[
        width=0.9\linewidth,  % Set to full width of the column
        height=0.9\linewidth,  % Increased height for more stretch
        grid=both,
        xlabel={x [m]},
        ylabel={y [m]},
        legend style={
            at={(1,1)}, 
            anchor=north east,  % Top right corner of the plot
            legend columns=1,  % Single column for the legend
            column sep=1ex,    % Space between columns
            legend cell align={left},
            thick,
            font=\scriptsize,  % Make legend labels smaller
            fill=none,  % No background for the legend
            draw=none,  % No border for the legend
        },
        tick label style={font=\small},
        label style={font=\small},
        axis line style={thick},
    ]

    % Ground Truth with Transparency and Reflective Look (only line, no points)
    \addplot[
        draw=black,
        line width=3pt,  % Make the line 4 times thicker
        dashed,  % Line connecting the points
        opacity=0.7,  % Make line 70% opaque
        samples=20  % Reduce to 20 samples from the data
    ] table {figs/pgfplots/pinc_data_gt.txt};
    \addlegendentry{Ground Truth}

    % PINC with ConFig (only line, no points)
    \addplot[
        draw=myred,  % Set to custom red color
        line width=3pt,  % Make the line 4 times thicker
        solid,  % Dashed line connecting the points
        opacity=0.6,  % Make line 60% opaque
        samples=20  % Reduce to 20 samples from the data
    ] table {figs/pgfplots/pinc_data_config.txt};
    \addlegendentry{PINC (with ConFig)}

    % PINC with Normalization (only line, no points)
    \addplot[
        draw=mygreen,  % Set to custom green color
        line width=3pt,  % Make the line 4 times thicker
        solid,  % Dotted line connecting the points
        opacity=0.7,  % Make line 70% opaque
        samples=20  % Reduce to 20 samples from the data
    ] table {figs/pgfplots/pinc_data_normalization.txt};
    \addlegendentry{PINC (with Normalization)}

    % Add starting point marker at (-0.2, -0.4) as 'X' shape and label it
    \addplot[
        only marks,
        mark=x,  % 'X' shaped marker
        mark options={scale=1.5, line width=2pt, fill=black},  % Custom marker size, line width and color
    ] coordinates {(-0.11,-0.4)};
    \node at (axis cs:-0.1,-0.4) [anchor=north, font=\small] {Initial Position};
    
    \end{axis}
    \end{tikzpicture}
    \caption{Many-step-ahead prediction position trajectories using the best model configuration with ConFIG and gradient normalization.}
      \label{fig:many_step_ahead_predictions}
\end{figure}

\subsection{Best Model Configuration Experiments}
\label{subsec:best_confs}
Based on our findings, we combined the optimal architectural and training choices, utilizing only data and physics losses. Additionally, we evaluated the impact of adding extra collocation points, which had previously improved Valid Prediction Times (VPTs).
These models were trained on a more complex dataset using a learning rate scheduler and a batch size of $N_B=10$ for $2400$ and $4800$ epochs. The performance metrics are presented in Fig. \ref{fig:final_results}. 
%With the learning rate scheduler, gradient normalization outperformed ConFIG, and adding extra collocation points did not enhance performance.
Thus, the best PINC configuration employs the data and physics losses, learning rate scheduling, gradient normalization, a batch size of $N_B=10$, and a single collocation point in the prediction interval $T$.
Fig.~\ref{fig:many_step_ahead_predictions} illustrates the performance differences between ConFIG and gradient normalization models.

%%%%%%%%%%%%%%%
% \begin{figure}[!t]
% z    \centering
%     \input{figs/pgfplots/computational_experiment}  % Reference to external plot file
%     \caption{\textcolor{red}{run models to get correct numbers  and refer in text} Comparison of computational time between approaches.}
%     \label{fig:comp_time}
% \end{figure}

%%%%%%%%%%%%%%%%%

% \begin{figure}[!b]
%     \centering
%     \includegraphics[width=\columnwidth]{figs/xy_predictions.pdf}
%     \caption{Many-step-ahead prediction position trajectories using the best model configuration with ConFIG and gradient normalization.}
%     \label{fig:many_step_ahead_predictions}
% \end{figure}

%%%%%%%%%%%%%%%%%%%%%%%%%%%%%%%%%%%%%%%%%%%%%%%%%%%%%%%%%%%%%%%%%%%%%%%%%%%%%%%%%%%%%%%%%%%%%%%%%%%%
%%%%%%%%%%%%%%%%%%%%%%%%%%%%%%%%%%%%%%%%%%%%%%%%%%%%%%%%%%%%%%%%%%%%%%%%%%%%%%%%%%%%%%%%%%%%%%%%%%%%
%%%%%%%%%%%%%%%%%%%%%%%%%%%%%%%%%%%%%%%%%%%%%%%%%%%%%%%%%%%%%%%%%%%%%%%%%%%%%%%%%%%%%%%%%%%%%%%%%%%%

\section{Conclusions and Future Work} % ~0.25 pages
\label{sec:conclusions}
This work applied PINC to model an underwater vehicle's dynamics, mapping the current state, input, and time to the predicted next state at that given time instant. Empirically, combining one-step-ahead data loss with physics-based regularization provided the best long-horizon accuracy at minimal computational overhead. A key insight was that residual connections, treating the network output as an integral increment, are essential. Adaptive learning rate scheduling, gradient normalization and weighting, and batch sizes in the range of 10–20 all enhanced predictive performance.

%Despite these promising results, several refinements remain. A single fixed collocation point consistently offered strong results, suggesting alternative architectures such as RNNs could further reduce overhead.

Despite these promising results, several refinements remain. Training PINC models on real-world ROV trajectories would enable more realistic evaluation, potentially revealing limitations not captured in simulation. Furthermore, extending PINC to handle full rotational dynamics -- optimally representing roll, pitch, and yaw -- could improve accuracy for realistic manoeuvres. Finally, investigating the minimal set of losses, real-world testing, integration with MPC, and online adaptation would make PINC an even more versatile framework for ROV applications.

%%%%%%%%%%%%%%%%%%%%%%%%%%%%%%%%%%%%%%%%%%%%%%%%%%%%%%%%%%%%%%%%%%%%%%%%%%%%%%%%
%%%%%%%%%%%%%%%%%%%%%%%%%%%%%%%%%%%%%%%%%%%%%%%%%%%%%%%%%%%%%%%%%%%%%%%%%%%%%%%%
%%%%%%%%%%%%%%%%%%%%%%%%%%%%%%%%%%%%%%%%%%%%%%%%%%%%%%%%%%%%%%%%%%%%%%%%%%%%%%%%
\balance
\bibliographystyle{IEEEtran}
\bibliography{bibliography}

% Generated by IEEEtran.bst, version: 1.14 (2015/08/26)
\begin{thebibliography}{10}
\providecommand{\url}[1]{#1}
\csname url@samestyle\endcsname
\providecommand{\newblock}{\relax}
\providecommand{\bibinfo}[2]{#2}
\providecommand{\BIBentrySTDinterwordspacing}{\spaceskip=0pt\relax}
\providecommand{\BIBentryALTinterwordstretchfactor}{4}
\providecommand{\BIBentryALTinterwordspacing}{\spaceskip=\fontdimen2\font plus
\BIBentryALTinterwordstretchfactor\fontdimen3\font minus \fontdimen4\font\relax}
\providecommand{\BIBforeignlanguage}[2]{{%
\expandafter\ifx\csname l@#1\endcsname\relax
\typeout{** WARNING: IEEEtran.bst: No hyphenation pattern has been}%
\typeout{** loaded for the language `#1'. Using the pattern for}%
\typeout{** the default language instead.}%
\else
\language=\csname l@#1\endcsname
\fi
#2}}
\providecommand{\BIBdecl}{\relax}
\BIBdecl

\bibitem{amer2023unav}
A.~Amer, O.~{\'A}lvarez-Tu{\~n}{\'o}n, H.~{\.I}. U{\u{g}}urlu, J.~L.~F. Sejersen, Y.~Brodskiy, and E.~Kayacan, ``Unav-sim: A visually realistic underwater robotics simulator and synthetic data-generation framework,'' in \emph{2023 21st International Conference on Advanced Robotics (ICAR)}, 2023, pp. 570--576.

\bibitem{kunz2008deep}
C.~Kunz, C.~Murphy, R.~Camilli, H.~Singh, J.~Bailey, R.~Eustice, M.~Jakuba, K.-i. Nakamura, C.~Roman, T.~Sato \emph{et~al.}, ``Deep sea underwater robotic exploration in the ice-covered arctic ocean with auvs,'' in \emph{2008 IEEE/RSJ International Conference on Intelligent Robots and Systems}.\hskip 1em plus 0.5em minus 0.4em\relax IEEE, 2008, pp. 3654--3660.

\bibitem{amer2023visual}
A.~Amer, M.~Mehndiratta, J.~le~Fevre~Sejersen, H.~X. Pham, and E.~Kayacan, ``Visual tracking nonlinear model predictive control method for autonomous wind turbine inspection,'' in \emph{2023 21st International Conference on Advanced Robotics (ICAR)}.\hskip 1em plus 0.5em minus 0.4em\relax IEEE, 2023, pp. 431--438.

\bibitem{lakshminarayanan2024estimation}
S.~Lakshminarayanan, D.~Duecker, A.~Sarabakha, A.~Ganguly, L.~Takayama, and S.~Haddadin, ``{Estimation of External Force Acting on Underwater Robots},'' in \emph{2024 IEEE 20th International Conference on Automation Science and Engineering~(CASE)}, 2024, pp. 3125--3131.

\bibitem{karniadakis2021physics}
G.~E. Karniadakis, I.~G. Kevrekidis, L.~Lu, P.~Perdikaris, S.~Wang, and L.~Yang, ``Physics-informed machine learning,'' \emph{Nature Reviews Physics}, vol.~3, no.~6, pp. 422--440, 2021.

\bibitem{drgona2025safe}
J.~Drgona, T.~X. Nghiem, T.~Beckers, M.~Fazlyab, E.~Mallada, C.~Jones, D.~Vrabie, S.~L. Brunton, and R.~Findeisen, ``Safe physics-informed machine learning for dynamics and control,'' \emph{arXiv preprint arXiv:2504.12952}, 2025.

\bibitem{antonelo_physics-informed_2024}
E.~A. Antonelo, E.~Camponogara, L.~O. Seman, E.~R. de~Souza, J.~P. Jordanou, and J.~F. Hubner, ``\BIBforeignlanguage{en}{Physics-{Informed} {Neural} {Nets} for {Control} of {Dynamical} {Systems}},'' \emph{\BIBforeignlanguage{en}{Neurocomputing}}, vol. 579, Apr. 2024, arXiv:2104.02556 [cs].

\bibitem{brunton_data-driven_2021}
S.~L. Brunton and J.~N. Kutz, \emph{\BIBforeignlanguage{en}{Data-{Driven} {Science} and {Engineering}}}.\hskip 1em plus 0.5em minus 0.4em\relax Cambridge University Press, 2021.

\bibitem{amer2025}
A.~Amer, M.~Mehndiratta, Y.~Brodskiy, and E.~Kayacan, ``Empowering autonomous underwater vehicles using learning-based model predictive control with dynamic forgetting gaussian processes,'' \emph{IEEE Transactions on Control Systems Technology}, 2025.

\bibitem{liang2025adaptive}
W.~Liang, A.~Amer, M.~Mehndiratta, Z.~Chen, B.~Yao, and E.~Kayacan, ``Adaptive robust control integrated with gaussian processes for quadrotors: Enhanced accuracy, fault tolerance and anti-disturbance,'' \emph{IEEE Transactions on Systems, Man, and Cybernetics: Systems}, 2025.

\bibitem{faria_data-driven_2024}
R.~Faria, B.~Capron, A.~Secchi, and M.~De~Souza, ``\BIBforeignlanguage{en}{A data-driven tracking control framework using physics-informed neural networks and deep reinforcement learning for dynamical systems},'' \emph{\BIBforeignlanguage{en}{Engineering Applications of Artificial Intelligence}}, vol. 127, Jan. 2024.

\bibitem{ramp}
S.~Sanyal and K.~Roy, ``Ramp-net: A robust adaptive mpc for quadrotors via physics-informed neural network,'' in \emph{2023 IEEE International Conference on Robotics and Automation (ICRA)}.\hskip 1em plus 0.5em minus 0.4em\relax IEEE, 2023, pp. 1019--1025.

\bibitem{lutter_combining_2023}
M.~Lutter and J.~Peters, ``\BIBforeignlanguage{en}{Combining physics and deep learning to learn continuous-time dynamics models},'' \emph{\BIBforeignlanguage{en}{The International Journal of Robotics Research}}, vol.~42, no.~3, pp. 83--107, Mar. 2023.

\bibitem{ma_development_2024}
J.~Ma, Y.~Li, J.~Tu, Y.~Zhang, J.~Ai, and Y.~Dong, ``\BIBforeignlanguage{en}{Development and {Implementation} of {Physics}-{Informed} {Neural} {ODE} for {Dynamics} {Modeling} of a {Fixed}-{Wing} {Aircraft} {Under} {Icing}/{Fault}},'' \emph{\BIBforeignlanguage{en}{Guidance, Navigation and Control}}, vol.~04, no.~01, Feb. 2024.

\bibitem{chen2018neural}
R.~T. Chen, Y.~Rubanova, J.~Bettencourt, and D.~K. Duvenaud, ``Neural ordinary differential equations,'' \emph{Advances in neural information processing systems}, vol.~31, 2018.

\bibitem{zhao_research_2024}
Y.~Zhao, Z.~Hu, W.~Du, L.~Geng, and Y.~Yang, ``Research on {Modeling} {Method} of {Autonomous} {Underwater} {Vehicle} {Based} on a {Physics}-{Informed} {Neural} {Network},'' \emph{Journal of Marine Science and Engineering}, vol.~12, no.~5, p. 801, May 2024.

\bibitem{gao_sim--real_2024}
J.~Gao, M.~Y. Michelis, A.~Spielberg, and R.~K. Katzschmann, ``Sim-to-{Real} of {Soft} {Robots} with {Learned} {Residual} {Physics},'' \emph{IEEE Robotics and Automation Letters}, pp. 1--8, 2024, conference Name: IEEE Robotics and Automation Letters.

\bibitem{krauss_domain-decoupled_2024}
H.~Krauss, T.-L. Habich, M.~Bartholdt, T.~Seel, and M.~Schappler, ``Domain-decoupled {Physics}-informed {Neural} {Networks} with {Closed}-form {Gradients} for {Fast} {Model} {Learning} of {Dynamical} {Systems},'' Aug. 2024.

\bibitem{kittelsen_physics-informed_2024}
J.~E. Kittelsen, E.~A. Antonelo, E.~Camponogara, and L.~S. Imsland, ``Physics-{Informed} {Neural} {Networks} with {Skip} {Connections} for {Modeling} and {Control} of {Gas}-{Lifted} {Oil} {Wells},'' \emph{Applied Soft Computing}, vol. 158, p. 111603, Jun. 2024.

\bibitem{liu_config_2024}
Q.~Liu, M.~Chu, and N.~Thuerey, ``{ConFIG}: {Towards} {Conflict}-free {Training} of {Physics} {Informed} {Neural} {Networks},'' Aug. 2024.

\bibitem{nicodemus_physics-informed_2022}
J.~Nicodemus, J.~Kneifl, J.~Fehr, and B.~Unger, ``Physics-informed {Neural} {Networks}-based {Model} {Predictive} {Control} for {Multi}-link {Manipulators},'' \emph{IFAC-PapersOnLine}, vol.~55, no.~20, pp. 331--336, Jan. 2022.

\bibitem{fossen}
T.~I. Fossen, ``\BIBforeignlanguage{eng}{Handbook of marine craft hydrodynamics and motion control},'' in \emph{\BIBforeignlanguage{eng}{Handbook of marine craft hydrodynamics and motion control}}, 2nd~ed.\hskip 1em plus 0.5em minus 0.4em\relax Hoboken, NJ: Wiley, 2021.

\bibitem{raissi_physics-informed_2019}
M.~Raissi, P.~Perdikaris, and G.~E. Karniadakis, ``Physics-informed neural networks: {A} deep learning framework for solving forward and inverse problems involving nonlinear partial differential equations,'' \emph{Journal of Computational Physics}, vol. 378, pp. 686--707, Feb. 2019.

\bibitem{ba_layer_2016}
J.~L. Ba, J.~R. Kiros, and G.~E. Hinton, ``Layer {Normalization},'' Jul. 2016.

\end{thebibliography}

\end{document}